\documentclass[lettersize,journal,twoside]{IEEEtran}
\usepackage{amsmath,amsfonts,amssymb}
\usepackage{algorithmic}
\usepackage{algorithm}
\usepackage{array}
\usepackage[caption=false,font=normalsize,labelfont=sf,textfont=sf]{subfig}
\usepackage{textcomp}
\usepackage{stfloats}
\usepackage{url}
\usepackage{verbatim}
\usepackage{graphicx}
\usepackage{cite} 
\usepackage{booktabs}
\usepackage{xcolor} 
\usepackage{float}
\usepackage{hyperref}
\usepackage{siunitx}
\usepackage{makecell}   

\begin{document} 

\title{FlexiCup: Wireless Multimodal Suction Cup with Dual-Zone Vision-Tactile Sensing}
\author{Junhao Gong$^{1,2*}$, Shoujie Li$^{1*\dagger}$, Kit-Wa Sou$^{1}$, Changqing Guo$^{1}$, Hourong Huang$^{1}$, Tong Wu$^{1}$, Yifan Xie$^{1}$, Chenxin Liang$^{1}$, Chuqiao Lyu$^{1}$, Xiaojun Liang$^{2}$, Wenbo Ding$^{1\dagger}$%
\thanks{Manuscript received: November 28, 2025; Revised: February 12, 2026; Accepted: March 17, 2026. This paper was recommended for publication by Editor Ashis Banerjee upon evaluation of the Associate Editor and Reviewers' comments.}%
\thanks{This work was supported by National Key R\&D Program of China (No.2024YFB3816000), Guangdong Innovative and Entrepreneurial Research Team Program (2021ZT09L197), and Shenzhen Science and Technology Program (No. KJZD20240903100905008), and Meituan Academy of Robotics Shenzhen, and Tsinghua University Meituan Joint Institute for Digital Life.}%
\thanks{*These authors contributed equally to this work.}%
\thanks{$^{\dagger}$Corresponding authors: Shoujie Li (lsj20@mails.tsinghua.edu.cn), Wenbo Ding (ding.wenbo@sz.tsinghua.edu.cn)}%
\thanks{$^{1}$Shenzhen Ubiquitous Data Enabling Key Lab, Shenzhen International Graduate School, Tsinghua University, Shenzhen 518055, China.}%
\thanks{$^{2}$Peng Cheng Laboratory, Shenzhen 518055, China.}%
\thanks{Digital Object Identifier (DOI): see top of this page.}%
} 
\markboth{IEEE ROBOTICS AND AUTOMATION LETTERS. PREPRINT VERSION. ACCEPTED MARCH, 2026}%
{Gong \MakeLowercase{\textit{et al.}}: FlexiCup: Wireless Multimodal Suction Cup with Dual-Zone Vision-Tactile Sensing}

\maketitle

\begin{abstract}
Conventional suction cups lack sensing capabilities for contact-aware manipulation in unstructured environments.
This paper presents FlexiCup, a multimodal suction cup with wireless electronics that integrate dual-zone vision-tactile sensing.
The central zone dynamically switches between vision and tactile modalities via illumination control, while the peripheral zone provides continuous spatial awareness.
The modular mechanical design supports both vacuum (sustained-contact adhesion) and Bernoulli (contactless lifting) actuation while maintaining the identical dual-zone sensing architecture, demonstrating sensing-actuation decoupling where sensing and actuation principles are orthogonally separable.
We validate hardware versatility through dual control paradigms.
Modular perception-driven grasping achieves comparable success rates across vacuum (90.0\%) and Bernoulli (86.7\%) modes using identical sensing and control pipelines, validating the sensing architecture's effectiveness across fundamentally different pneumatic principles.
Diffusion-based end-to-end learning achieves 73.3\% and 66.7\% success on contact-aware manipulation tasks, with ablation studies confirming 13\% improvements from multi-head attention coordinating dual-zone observations.
Hardware designs, firmware, and experimental videos are available at the companion website: \url{https://flexicup.junhaogong.top}.
\end{abstract} 
\begin{IEEEkeywords}
Multimodal sensing, Suction manipulation, Vision-tactile sensing, Policy learning
\end{IEEEkeywords}

\IEEEpeerreviewmaketitle

\section{INTRODUCTION}

\IEEEPARstart{S}{uction} cups have been widely adopted in robotic manipulation for handling diverse object geometries~\cite{Huh2021, Lee2024, Davis2008, Liu2020, Yoo2023}. Effective manipulation in unstructured environments requires target identification, obstacle detection, contact verification, and surface adaptation. Traditional suction systems rely on preprogrammed trajectories without sensing feedback~\cite{Doi2020, Aoyagi2020}.

\begin{figure}[t]
    \centering
    \includegraphics[width=\columnwidth]{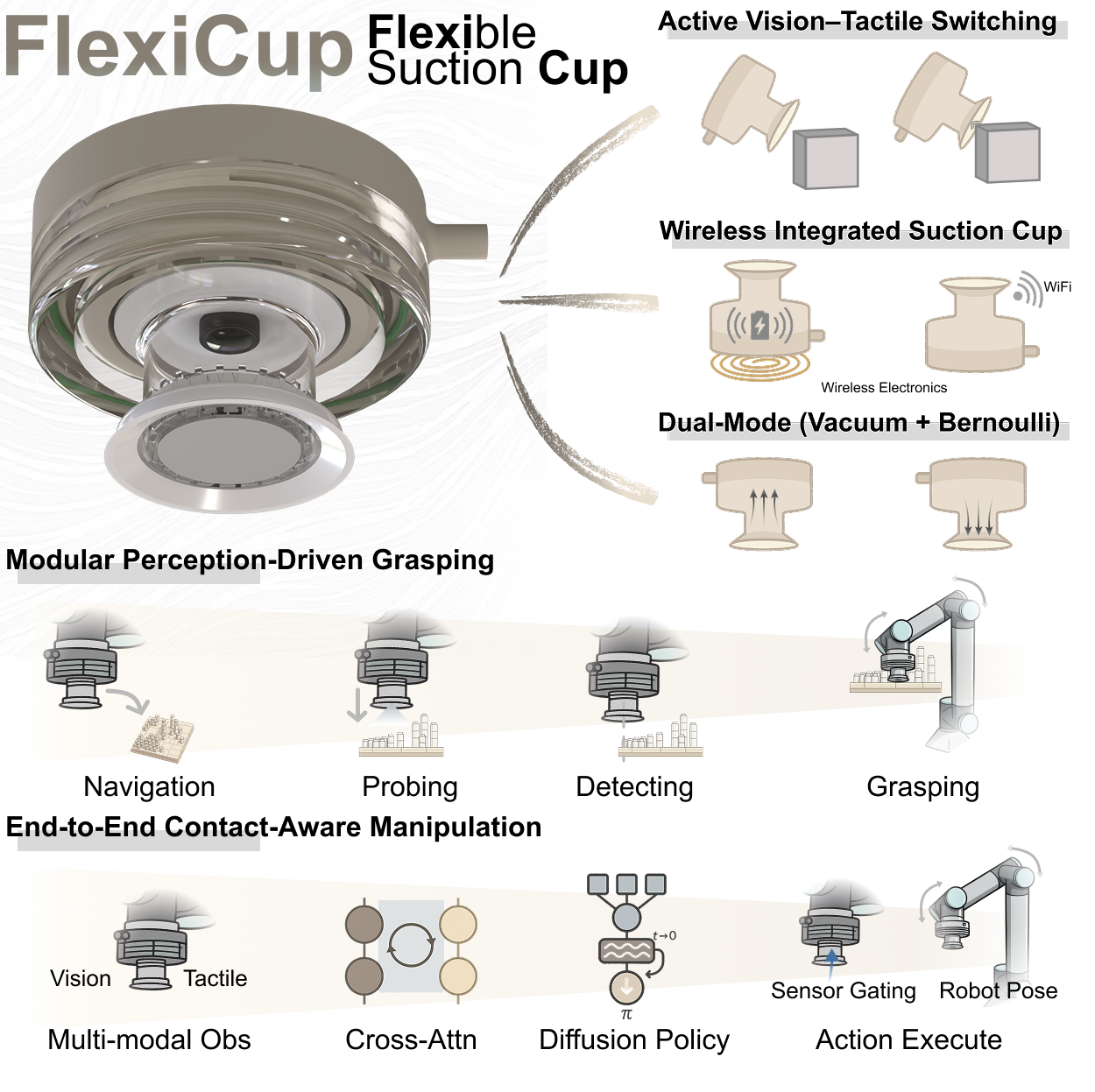}
    \caption{FlexiCup overview showing hardware design (top), modular grasping (middle), and end-to-end learning (bottom).}
    \label{fig:FigCover}
\end{figure} 

Recent efforts have integrated sensing into suction cups to enable contact-aware manipulation. Camera-based tactile systems~\cite{vanVeggel2025, Yuan2025} achieve high-resolution contact imaging but dedicate their entire optical field to the tactile membrane, sacrificing peripheral spatial awareness. Systems using external cameras~\cite{Lee2024, wang2025demonstratingmultisuctionitempicking} can observe the workspace but suffer from occlusion during contact.
These architectural constraints force a trade-off: existing sensory suction systems must choose between high-resolution contact sensing and spatial context awareness.

The visuotactile sensing community has demonstrated that coordinating vision and tactile observations enables robust manipulation under varied conditions~\cite{Zhao2025, Jones2025, Zhao2024TactileSensing, Lin2025}. However, these systems typically employ gripper-based designs~\cite{Wang2024, DelBianco2024} that require opposing contact surfaces or graspable features, limiting their applicability to scenarios such as handling flat featureless objects or accessing confined spaces with restricted side clearance.

Motivated by these advances in multimodal sensing, we present FlexiCup, a wireless multimodal suction cup that addresses the architectural limitations in existing sensory suction systems through dual-zone vision-tactile sensing while operating in manipulation scenarios where gripper-based approaches face geometric constraints (Fig.~1). A key insight is decoupling sensing from actuation: the dual-zone vision-tactile architecture operates across both vacuum and Bernoulli pneumatic principles. The architecture implements dual-zone sensing: a central zone switches between vision and tactile modalities via illumination control, while a peripheral zone maintains continuous spatial awareness.

This architecture enables adaptive manipulation requiring coordinated visual search, obstacle avoidance, and contact verification—capabilities that existing sensory suction systems cannot simultaneously provide. By overcoming the morphological constraints of gripper-based counterparts on flat or confined targets, FlexiCup validates its efficacy through experiments on varying obstacle densities, confined space extraction, and multi-phase manipulation sequences requiring continuous feedback loop.

The key contributions of this work include:
\begin{itemize}
\item Sensing-actuation decoupling through modular design, enabling identical sensing and control pipelines across vacuum and Bernoulli modes.

\item Dual-zone vision-tactile architecture with wireless electronics, implementing illumination-controlled modality switching between contact imaging and spatial context.

\item Dual-paradigm validation combining modular grasping (90.0\% vacuum, 86.7\% Bernoulli) and diffusion-based learning with multi-head attention, confirmed by ablation studies and baseline comparisons.
\end{itemize}

\begin{table*}[htbp] 
    \centering
    \caption{Comparison of Representative Suction Cup Perception Systems}
    \label{tab:ComparisonSystems}
    \begin{tabular}{@{}lllllll@{}}
    \toprule
    \textbf{Work} & \textbf{Mechanism} & \textbf{Modality} & \textbf{Perception Type} & \textbf{Force} & \multicolumn{1}{c}{\textbf{Method}} & \textbf{Deployment}\\
    \midrule
    Doi et al. \cite{Doi2020} & Vacuum& Tactile & Capacitive  & --- & Stroke-based State Decision & Wired \\
    Aoyagi et al. \cite{Aoyagi2020} & Vacuum & Tactile & Piezoresistive & --- & Multi-cup Force Reasoning & Wired \\
    Lee et al. \cite{Lee2024} & Vacuum& Tactile & Pressure-based & 19 N & Model-based Haptic Search & Wired \\
    Shahabi et al. \cite{Shahabi2023} & Vacuum& Tactile & Piezoresistive & --- & ML-based Property Regression & Wired \\
    van Veggel et al. \cite{vanVeggel2025} & Vacuum& Tactile & \textbf{Vision-based} & 9.35 N & CNN-based Pose Correction & Wired \\
    Yue et al. \cite{Yue2025} & Vacuum & Tactile & Pressure-based & --- & Hierarchical Embodied Control & Wired \\
    Jang et al. \cite{Jang2021} & Vacuum & Tactile & Pressure-based & 2.8 N & Closed-loop Force Regulation & Wired \\ 
    \textbf{Ours} & \textbf{Vacuum / Bernoulli} & \textbf{Vision + Tactile} & \textbf{Vision-based} & \textbf{41.5 N} &
    \makecell[l]{\textbf{Modular Perception Driven} \\ / \textbf{Diffusion-based Imitation Learning}} &
    \textbf{Wireless} \\
    
    \bottomrule
    \end{tabular}
    \end{table*}

\section{Related Work}

Suction manipulation has advanced through diverse sensing modalities (Table~\ref{tab:ComparisonSystems}).
Early systems employed capacitive~\cite{Doi2020} and piezoresistive~\cite{Aoyagi2020,Shahabi2023} sensing for basic contact detection, while pressure-based approaches~\cite{Lee2024,Yue2025,Jang2021} enabled model-based haptic search and force regulation.
Recent camera-based tactile sensors~\cite{vanVeggel2025,Shiratori2024}, building on foundational vision-based tactile technologies~\cite{gelsight, digit, gelslim}, achieve high-resolution deformation imaging but dedicate their entire optical field to contact sensing, precluding simultaneous spatial awareness.
Integration challenges remain: electronic tactile arrays can compromise pneumatic sealing, while maintaining perception throughout contact-to-non-contact transitions requires architectural solutions beyond single-modality designs.

Control paradigms have similarly evolved from classical planning~\cite{Lee2024, Motoda2024} to learning-based methods~\cite{Shahabi2023,vanVeggel2025,TactileRL2024,song2025overview,An2025}, yet deployment constraints persist—existing systems require wired connections for high-bandwidth sensing.
This work addresses these limitations through wireless electronics, dual-zone vision-tactile architecture enabling modality switching, and modular dual-mode design supporting both classical and diffusion-based control while achieving 41.5 N force capacity.

\begin{figure*}[ht]
    \centering
    \includegraphics[width=2\columnwidth]{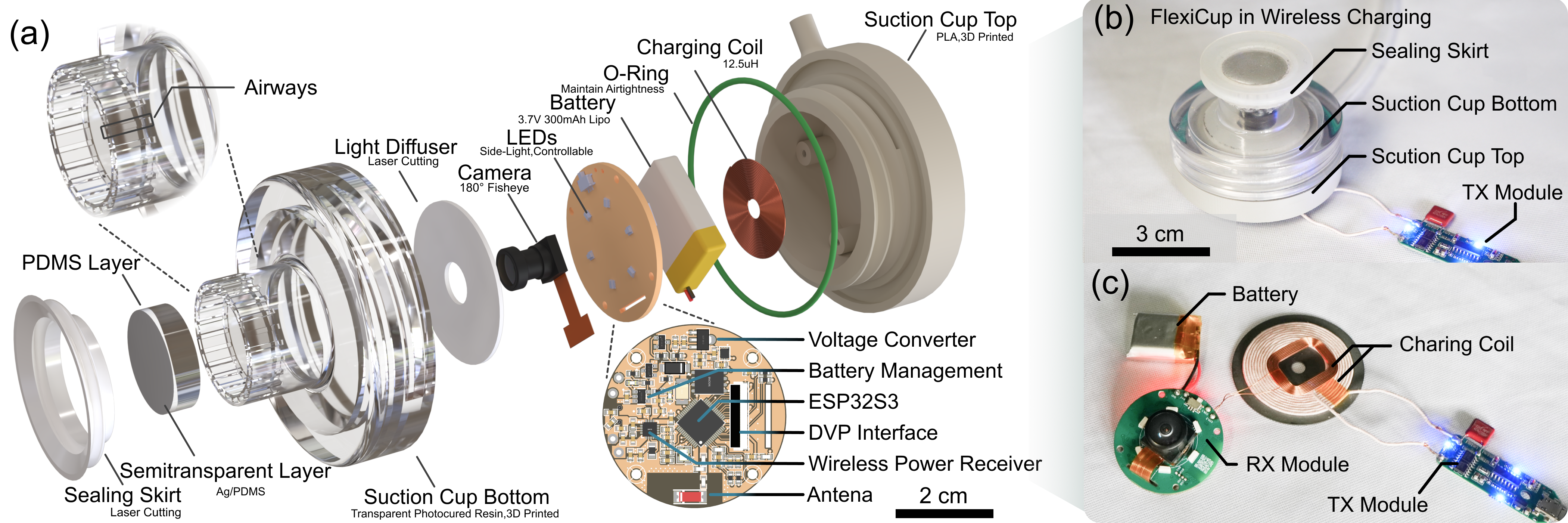}
    \caption{System architecture showing (a) modular component integration, (b) wireless charging configuration, and (c) standalone charging module.}
    \label{fig:FigStructure}
\end{figure*}

\section{Hardware Design}

The FlexiCup hardware addresses the trade-off between contact imaging and spatial awareness by realizing dual-zone vision–tactile sensing in a wireless, modular suction morphology. It implements a layered pneumatic–optical–electronic stack and interchangeable bottom housings that decouple sensing from actuation while remaining compatible with both vacuum and Bernoulli principles; the following subsections detail the guiding design principles and system architecture.

\subsection{Design Principles}

FlexiCup integrates vision and tactile modalities within a unified optical framework, employs wireless electronics to eliminate electrical tethering, and implements modular mechanical design enabling dual-mode operation with complementary actuation mechanisms—vacuum for contact-based adhesion and Bernoulli for contactless lifting—while maintaining identical core electronics and dual-zone sensing architecture.

\subsection{System Architecture and Integration}

The system adopts a modular layered architecture integrating pneumatic actuation, electronic control, and optical sensing (Fig.~\ref{fig:FigStructure}(a)).
The modular design separates task-specific pneumatic actuation in the bottom housing from shared electronic control and optical sensing in the top layer, enabling rapid reconfiguration between suction modes.

The bottom housing provides the pneumatic interface and contact surface, with a central groove accommodating the PDMS membrane for sealing and sensing. Pneumatic airways route around the membrane perimeter, physically separating actuation from sensing. The airway geometry differs between Vacuum and Bernoulli configurations, while the membrane mounting interface remains consistent. The bottom housing engages with the top through threaded connection with fluoroelastomer O-ring seal.

The top housing mounts a PCB assembly centered on an ESP32S3 microcontroller (3.7 V, 300 mAh battery, 12.5 $\mu$H wireless charging coil), streaming 640×480 images at 30 Fps over Wi-Fi.
The battery provides approximately 30 minutes of continuous operation, with thermal management facilitated by pneumatic airflow and 2-hour wireless charging at 200 mA (Fig.~\ref{fig:FigStructure}(b,c)).
This electronic module remains identical across both suction modes, enabling wireless operation with electrical decoupling from the robot arm.

The top housing integrates an optical sensing system below the electronic module, comprising an OV5640 camera with 180° fisheye lens and LED arrays enabling dual-zone perception through the tactile membrane interface.

The pneumatic control system employs distinct configurations: vacuum mode utilizes a vacuum pump (750 W, 140 L/min) at -90 kPa maximum pressure, achieving 41.5 N normal force at -80 kPa, while Bernoulli mode relies on an air compressor (800 W, 65 L/min) with supply pressure up to 0.8 MPa. Pneumatic actuation is controlled wirelessly through solenoid valves triggered by the onboard microcontroller. The standardized interface enables rapid reconfiguration by exchanging the bottom housing and connecting to the corresponding pneumatic source.

\begin{figure}[!t] 
    \centering
    \includegraphics[width=1\columnwidth]{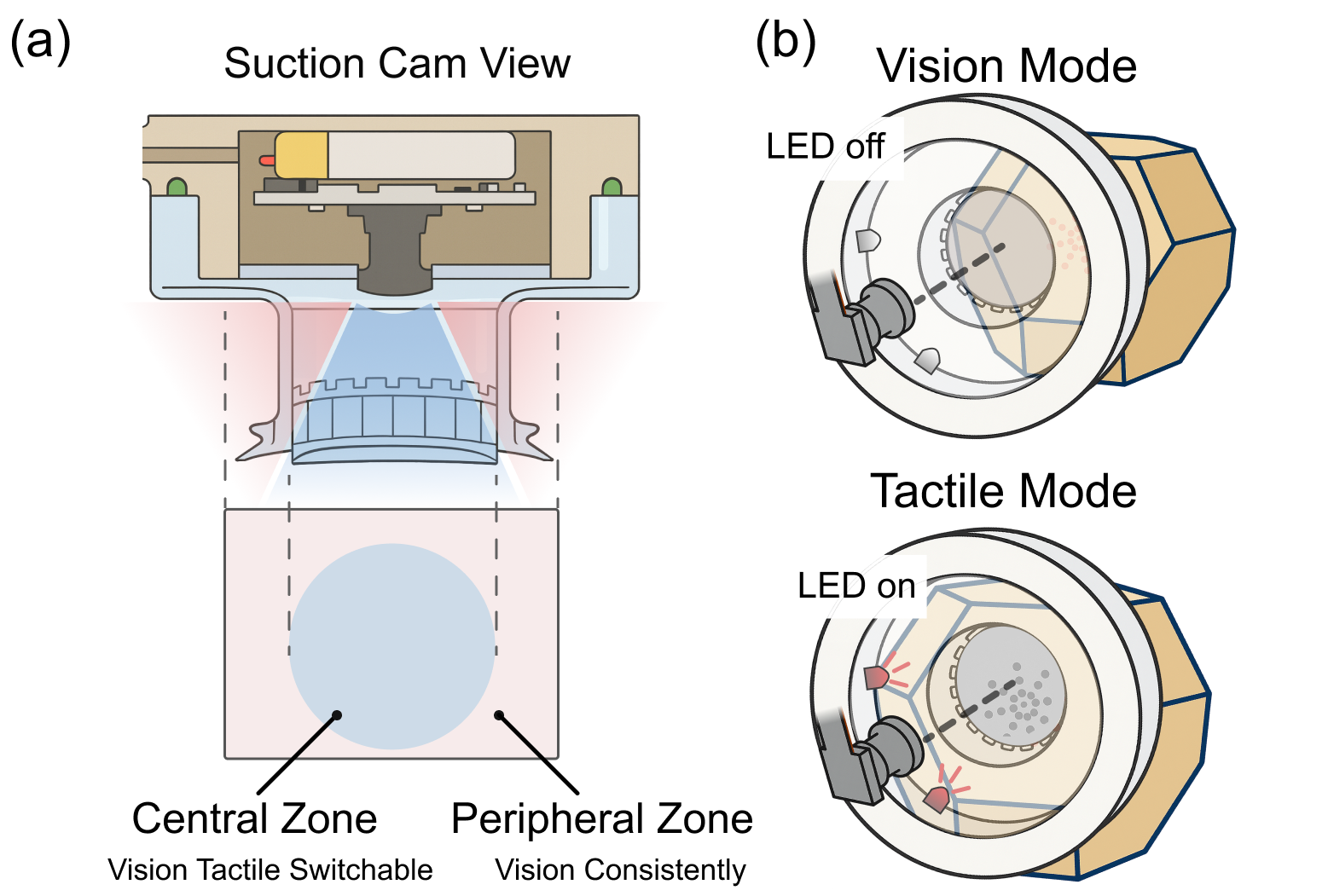}
    \caption{Vision-tactile sensing showing (a) dual-zone camera view and (b) modality switching via illumination control.}
    \label{fig:FigVisionTactileSensingSystem}
\end{figure}
 
\subsection{Vision-Tactile Sensing System}

The 180° fisheye camera captures two functional zones: the central zone enables switchable vision-tactile sensing, while the peripheral zone provides continuous visual awareness (Fig.~\ref{fig:FigVisionTactileSensingSystem}(a)).

The system switches modalities through illumination control in real-time by the ESP32S3 microcontroller with dynamic camera exposure and gain adjustments (Fig.~\ref{fig:FigVisionTactileSensingSystem}(b)).
In vision mode, LEDs remain inactive, allowing ambient light through the membrane for object detection.
In tactile mode, LEDs illuminate the membrane internally, imaging deformations induced by contact forces.
The semitransparent PDMS membrane exhibits ambient light sensitivity, addressed through high-intensity internal LEDs with optimized fisheye lens and camera exposure settings, though tactile features may become less pronounced under exceptionally strong environmental lighting.

\begin{figure}[!t]
    \centering
    \includegraphics[width=1\columnwidth]{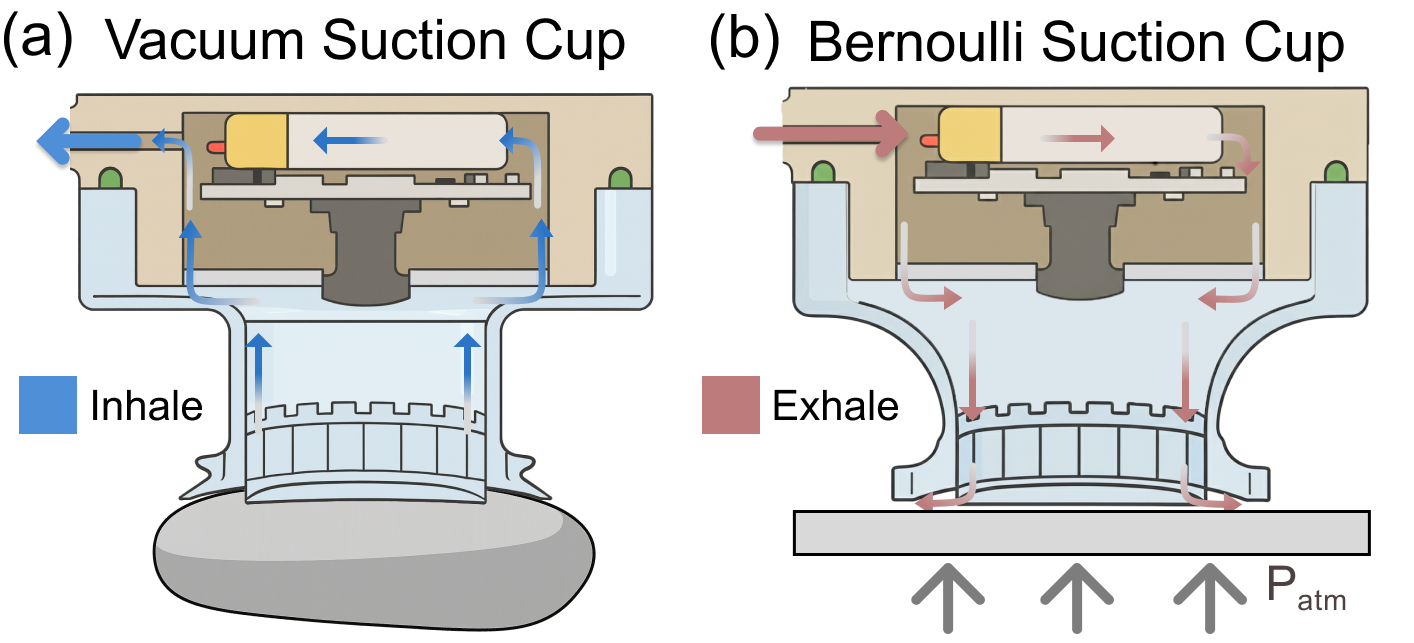}
    \caption{Dual-mode suction mechanisms demonstrating complementary actuation principles.
    (a) Vacuum mode generates adhesion through inward airflow requiring sustained contact.
    (b) Bernoulli mode creates lifting force through outward airflow enabling contactless lifting.
    Both modes share the identical dual-zone vision-tactile sensing architecture.
    }
    \label{fig:FigDualModeSuctionMechanisms}
\end{figure}

\subsection{Dual-Mode Suction Mechanisms with Complementary Sensing} 

Vacuum mode generates adhesion through negative pressure, requiring sustained contact that induces continuous membrane deformation, enabling dense tactile feedback throughout manipulation for deformable object handling and surface compliance detection (Fig.~\ref{fig:FigDualModeSuctionMechanisms}(a)).

Bernoulli mode creates contactless lifting through outward airflow (Fig.~\ref{fig:FigDualModeSuctionMechanisms}(b)), with the dual-zone architecture supporting visual perception and tactile verification during positioning.
Both modes share identical sensing pipelines, confirming sensing-actuation decoupling where mode selection depends on task requirements rather than sensing constraints.

\begin{figure}[!t]
    \centering
    \includegraphics[width=1\columnwidth]{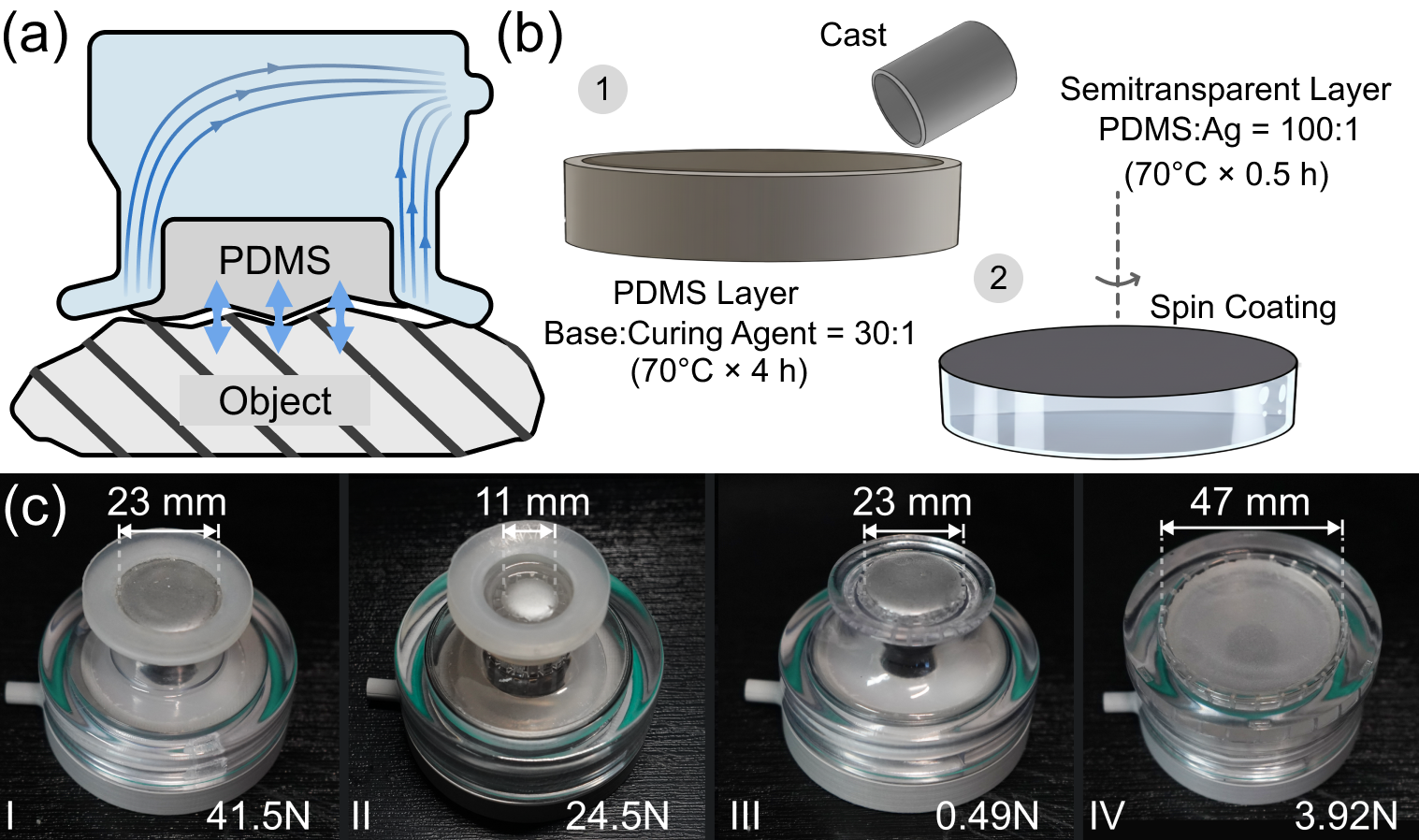}
    \caption{PDMS membrane design and fabrication. (a) PDMS membrane deformation during contact. (b) Dual-layer membrane fabrication process. (c) Four modular configurations (I-IV).}
    \label{fig:FigPDMSMembraneDesign}
\end{figure}

\subsection{PDMS Membrane Design}

The PDMS membrane provides tactile sensing and sealing through contact-induced deformation (Fig.~\ref{fig:FigPDMSMembraneDesign}(a)). The dual-layer system (Fig.~\ref{fig:FigPDMSMembraneDesign}(b)) employs a PDMS base layer (30:1 mass ratio, 70°C for 4 hours) providing compliance that captures surface details, and a semitransparent layer (Ag:PDMS 100:1, 70°C for 0.5 hours) providing the reflective surface for photometric tactile imaging. Four modular bottom configurations (I-IV, Fig.~\ref{fig:FigPDMSMembraneDesign}(c)) implement varying membrane diameters for vacuum operation. Configurations I-II are optimized for deformable objects and tactile sensitivity. Adhesion force ranges from sub-Newton to over 40 N across configurations. Configuration selection requires only bottom housing replacement while all sensing components remain unchanged.

\subsection{Force Characteristics from Vacuum Mode}

\begin{figure}[!t]
    \centering
    \includegraphics[width=1\columnwidth]{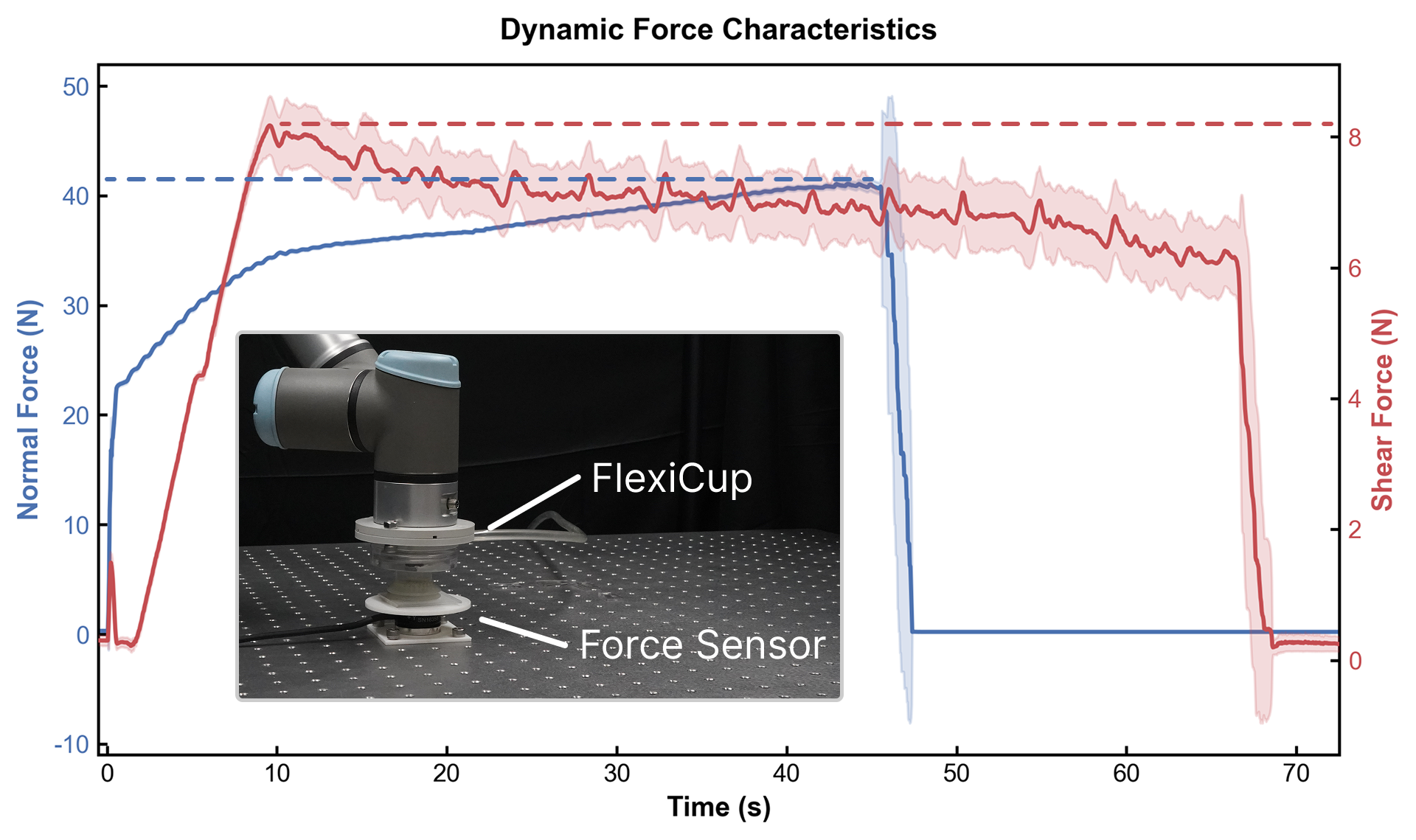}
    \caption{Force characteristics from vacuum mode. Blue curve: normal force evolution during vertical pull-off from contact to detachment. Red curve: shear force evolution during horizontal drag from contact to detachment. Inset: experimental setup with FlexiCup mounted on robot arm and force sensor fixed on platform.}
    \label{fig:FigForceTest}
\end{figure}

To characterize suction performance, we conducted automated force measurements using a 6-axis force/torque sensor on a smooth acrylic surface at -80 kPa vacuum pressure. The test protocol included normal pull-off tests measuring detachment force and horizontal drag tests measuring shear resistance, with each test repeated 20 times. The averaged force profiles (Fig.~\ref{fig:FigForceTest}) reveal transient behaviors during attachment and detachment. Results demonstrate mean maximum normal force of 41.5 N and shear force of 8.34 N, exceeding theoretical predictions (F = P × A $\approx$ 33.2 N) due to structural compliance increasing the effective sealing area.

\subsection{Multimodal Sensing Validation}

\begin{figure}[ht]
    \centering
    \includegraphics[width=1\columnwidth]{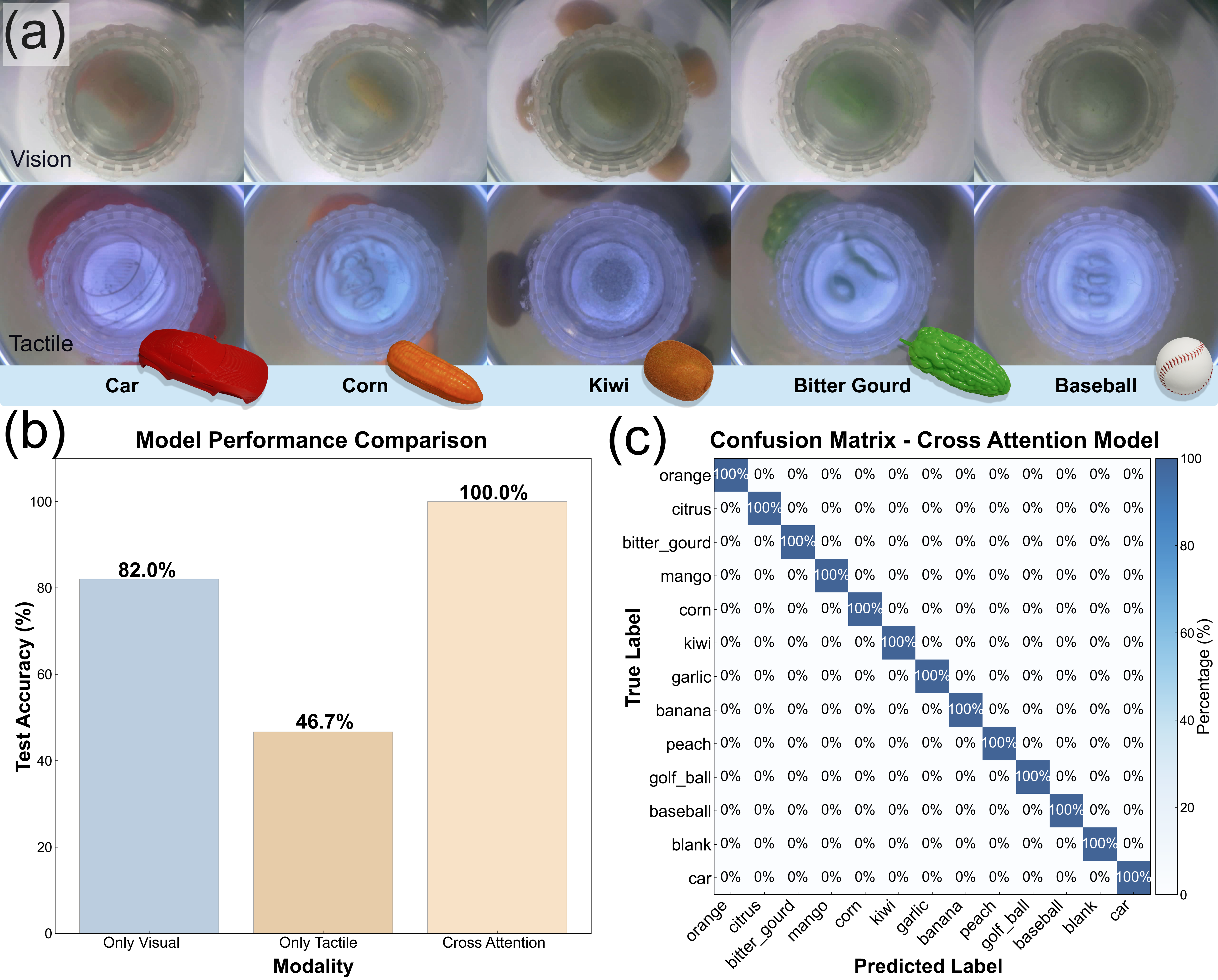}
    \caption{Multimodal recognition validation showing (a) vision-tactile sensing comparison across objects, (b) accuracy comparison (vision-only: 82.5\%, tactile-only: 46.7\%, multi-head attention: 100\%), and (c) confusion matrix.}
    \label{fig:FigVisualTactileModalitiesPhotos} 
\end{figure}

We conducted object recognition experiments across thirteen objects (Fig.~\ref{fig:FigVisualTactileModalitiesPhotos}(a)). Using ImageNet-pretrained ResNet-18, vision-only achieves 82.5\%, tactile-only 46.7\%, and multi-head attention fusion 100\% accuracy (Fig.~\ref{fig:FigVisualTactileModalitiesPhotos}(b)). The fusion model employs multi-head attention (8 heads, 512-d) trained on 1,300 paired samples across 13 categories with AdamW optimization. Results demonstrate that multimodal integration provides complementary information where vision captures global features and tactile reveals local contact details. The confusion matrix (Fig.~\ref{fig:FigVisualTactileModalitiesPhotos}(c)) confirms classification across all objects when both modalities are coordinated.

\begin{figure*}[ht]
    \centering
    \includegraphics[width=2\columnwidth]{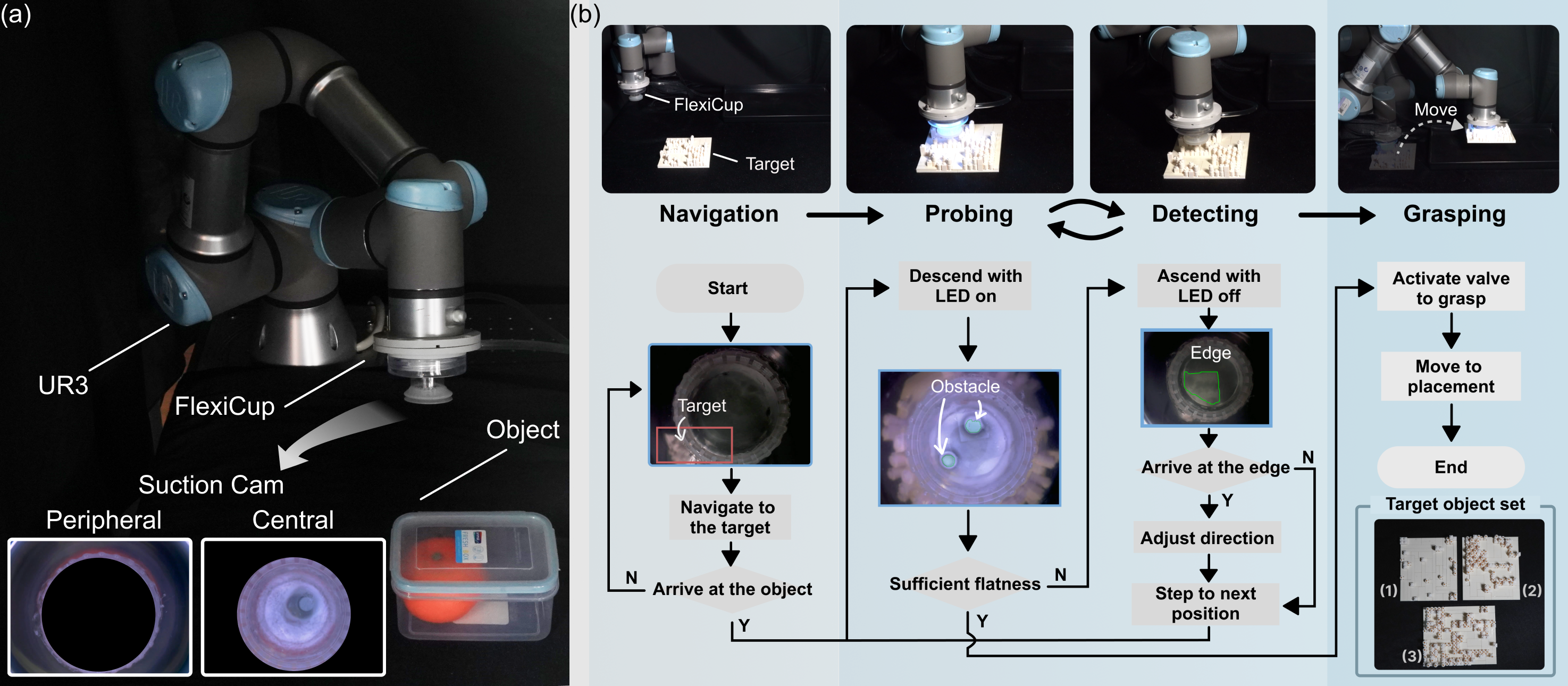}
    \caption{Experimental setup and modular control framework for perception-driven grasping. (a) FlexiCup mounted on UR3 robot. (b) Control pipeline integrating YOLOv8 and ResNet-34 for dual-mode validation. Target object set (bottom right): (1)-(3) with 25\%, 50\%, 75\% obstacle coverage.}
    \label{fig:FigExperimentSetup2}
\end{figure*}

\section{Modular Perception-Driven Grasping}

To validate that the dual-zone vision-tactile sensing architecture operates effectively
  across both vacuum and Bernoulli actuation principles, we conduct experiments using two
  control paradigms: modular perception-driven approach (this section) and learning-based
  policies (Section V). The modular approach demonstrates that both modes can leverage
  identical sensing pipelines and control logic with only pneumatic parameter adjustments,
  operating autonomously using FlexiCup's onboard dual-zone camera. All experiments use
  FlexiCup mounted on a UR3 robot arm (Fig.~\ref{fig:FigExperimentSetup2}(a)).

\subsection{Modular Control Framework}

The modular perception pipeline leverages YOLOv8n for peripheral target detection and ResNet-34 for central tactile verification, demonstrating hardware versatility across classical planning methods.
The YOLOv8n detector processes peripheral vision for target localization and workspace boundary detection.
When LED illumination is activated, the ResNet-34 segmentation model (pretrained on ImageNet) analyzes the central tactile imprint to identify contact regions and verify surface flatness.
With LED off, a YOLOv8n-seg model performs edge detection on the central visual stream to guide obstacle avoidance during spatial search.

As illustrated in Fig.~\ref{fig:FigExperimentSetup2}(b), the control pipeline integrates these modules with iterative feedback: peripheral vision detects the target and guides approach; upon arrival, tactile segmentation verifies contact and surface flatness; if unsuitable, the system transitions to vision mode and uses edge detection to guide spatial stepping; finally, valve activation initiates suction and transport.

\subsection{Experimental Configuration}

The validation employs identical sensing and control pipelines across both modes, with only bottom housing and pneumatic parameters varying.
The three perception modules were trained offline on task-specific datasets, achieving 87.7\% mAP50 for peripheral detection, 98.3\% IoU for tactile segmentation, and 99.5\% mAP50 for edge detection, all supporting real-time inference.

Test materials include custom LEGO boards with 25\%, 50\%, 75\% obstacle coverage.
We conduct 30 trials per mode (vacuum and Bernoulli), totaling 180 trials, with success criteria including navigation to target, obstacle-free region identification, stable attachment verified by tactile feedback, and transport without object loss.


\begin{table}[h]
    \centering
    \caption{Hardware Modality Validation: Success Rates Across Pneumatic Principles}
    \label{tab:hardware_modality_results}
    \begin{tabular*}{\columnwidth}{@{\extracolsep{\fill}}lcc}
    \toprule
    Test Material & Vacuum & Bernoulli \\
    \midrule
    (1) 25\% obstacle coverage & 90.0\% & 86.7\% \\
    (2) 50\% obstacle coverage & 93.3\% & 90.0\% \\
    (3) 75\% obstacle coverage & 86.7\% & 83.3\% \\
    \midrule
    \textbf{Mean} & \textbf{90.0\%} & \textbf{86.7\% }\\
    \bottomrule
    \end{tabular*}
\end{table}

\subsection{Results and Analysis}

Table~\ref{tab:hardware_modality_results} shows comparable success rates for vacuum (90.0\%) and Bernoulli (86.7\%), confirming sensing universality across contact-based and non-contact principles, control portability with only pneumatic parameter changes, and mechanical modularity through bottom-housing reconfiguration.

The comparable success rates validate sensing-actuation decoupling: both modes achieve effective manipulation using the shared dual-zone architecture, with performance differences attributable to actuation mechanisms rather than sensing limitations.

Failure cases for both modes occur when the discrete search strategy (1 cm step size) exhaustively traverses the workspace without identifying suitable attachment regions.
Success rates peak at moderate obstacle density (mean 91.7\%) compared to low and high densities (88.4\% and 85.0\%), as scattered obstacles can fragment surfaces into regions smaller than the suction cup footprint, affecting both vacuum and Bernoulli modes comparably.

Bernoulli's slightly lower success rate (86.7\% vs 90.0\%) stems from reduced adhesion force during the contactless lifting phase: while the shared sensing pipeline successfully identifies suitable regions, the weaker aerodynamic force occasionally fails to maintain grasp on challenging configurations where vacuum's contact-based adhesion succeeds.

However, Bernoulli provides unique capabilities: a complementary semiconductor wafer handling experiment demonstrates that vacuum-picked wafers exhibit visible contact smudging ($\sim$3.5 N peak contact force), while Bernoulli-picked wafers remain pristine. 
    F/T sensor measurements confirm that $F_z$ remained within 0.3~N throughout Bernoulli lifting---equivalent to the wafer's gravitational load---indicating zero mechanism-induced contact force, in contrast to the $\sim$3.5~N transient peak recorded under vacuum mode.
 Side-view footage on the companion website qualitatively corroborates the maintained air gap.

\begin{figure}[ht] 
    \centering
    \includegraphics[width=\columnwidth]{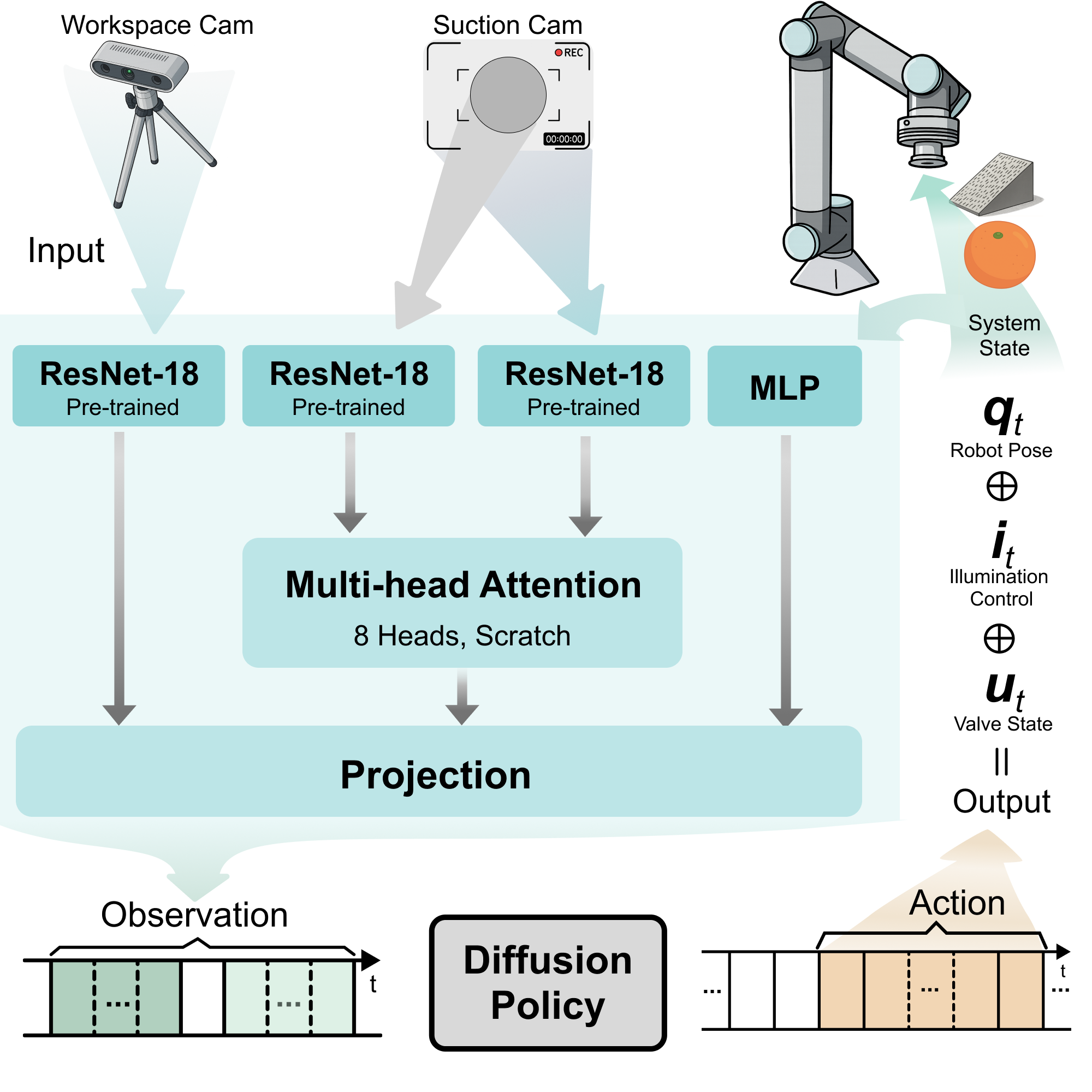}
    \caption{Diffusion policy framework combining workspace, peripheral, central, and state observations through parallel encoders and multi-head attention.}
    \label{fig:FigFramework}
\end{figure}

\section{End-to-End Contact-Aware Manipulation}

While the modular perception-driven approach validates hardware reconfigurability, it relies on discrete state transitions and manual threshold tuning. We develop a learning framework based on diffusion policies~\cite{Chi2023,Xue2025DemoGen} mapping multimodal observations to action trajectories.

\subsection{Multi-Modal Learning Framework}

We adopt the diffusion policy framework for sequential suction operations with vision-tactile modality transitions.

The learning experiments utilize vacuum mode (Configuration I) exclusively, as contact-aware policy learning requires continuous tactile feedback from sustained contact and rich membrane deformation. While Section IV demonstrated that the dual-zone sensing architecture operates effectively across both actuation principles, vacuum's sustained contact provides dense tactile gradients essential for learning fine-grained manipulation skills involving contact detection, surface compliance interpretation, and valve timing coordination. In contrast, Bernoulli's contactless lifting during suction minimizes membrane deformation, providing limited tactile information for learning contact dynamics despite supporting tactile verification during pre-suction positioning.

We extend the framework with multi-head attention to fuse dual-zone observations (Fig.~\ref{fig:FigFramework}), coordinating colocated heterogeneous sensory streams during vision-tactile switching.

\subsection{Multi-Modal Observation Encoding}

The system processes observations through specialized encoders.
The workspace view $I_t^{workspace}$ from the D435 camera provides global scene context, encoded via ImageNet pre-trained ResNet-18 to $f_t^{workspace} \in \mathbb{R}^{512}$. The dual-zone suction camera captures local observations: the central view $I_t^{central}$ switches between visual and tactile modes for contact sensing, and the peripheral view $I_t^{peripheral}$ maintains spatial awareness, both encoded through ResNet-18 to $f_t^{central}, f_t^{peripheral} \in \mathbb{R}^{512}$. The system state $s_t \in \mathbb{R}^8$ is processed through a 2-layer MLP to $f_t^{state}$.

\begin{figure*}[ht]
    \centering
    \includegraphics[width=2.0\columnwidth]{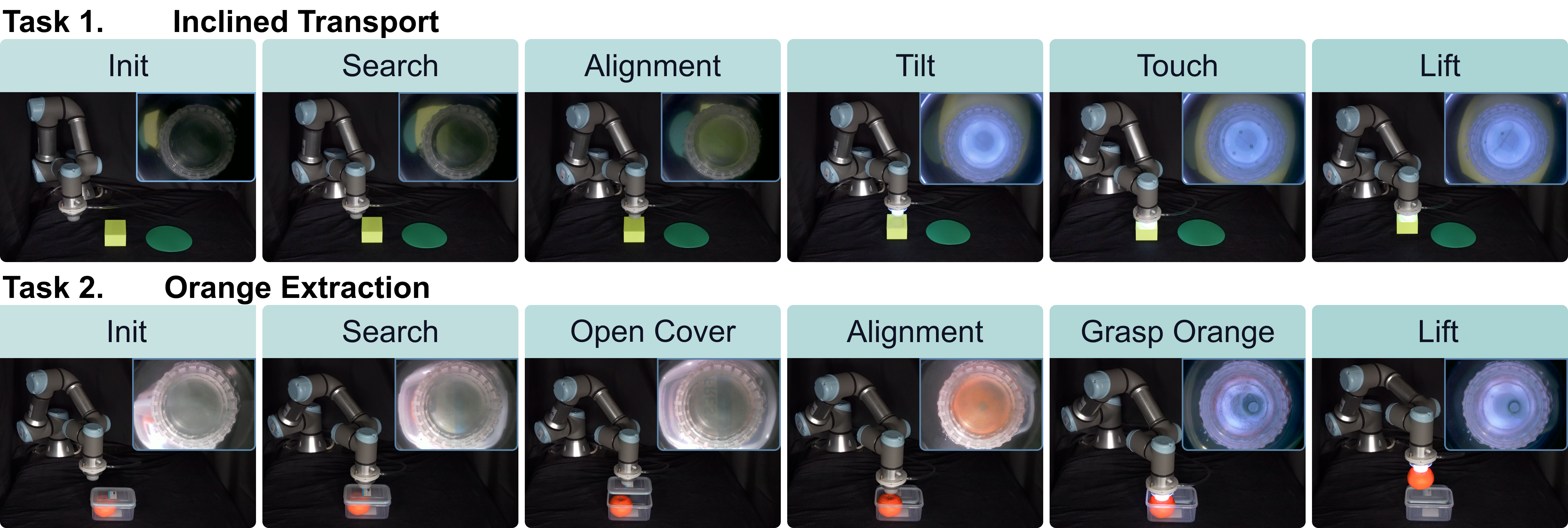}
    \caption{Experimental task demonstrations showing inclined transport and orange extraction sequences with multimodal sensing integration.}
    \label{fig:FigTaskRollouts}
\end{figure*}

\subsection{Feature Integration and Policy Learning}

The multi-head attention module (8 heads, 512 dimensions) coordinates the central and peripheral views, correlating contact details with spatial context during approach-to-manipulation transitions. Integration proceeds through two stages: workspace features combine with attended suction features, then concatenate with system state to form the complete observation representation. 

The diffusion policy generates action trajectories $a_t = [q_t, i_t, u_t]$ controlling robot joints ($q_t \in \mathbb{R}^6$), illumination switching ($i_t$), and pneumatic valve ($u_t$). The action chunking mechanism (8-step history, 48-step horizon) generates coordinated sequences for multi-phase operations.

\subsection{Experimental Evaluation}

We evaluate two tasks: inclined transport (150 demos) and orange extraction (100 demos), collected via kinesthetic teaching at 30 Hz and downsampled to 10 Hz, with randomized positions, angles, and orientations, and random cropping augmentation (76$\times$76 from 224$\times$224).
Inclined transport involves positioning above the surface (7 cm $\times$ 5.3 cm $\times$ 6 cm), searching for suitable contact regions, adjusting tilt angle guided by tactile feedback to match the inclined surface (5\textdegree, 10\textdegree, or 15\textdegree), verifying contact, and performing secure lifting.
Orange extraction consists of transparent cover removal (container: 15 cm $\times$ 10.5 cm $\times$ 7 cm) in vision mode, realignment above the orange (approximately 6 cm diameter), then tactile-guided grasping with LED-enabled contact detection.

Five configurations (Ours, w/o Multi-Head Attention, w/o Peripheral View, w/o Central View, Workspace Camera Only) and a BC-RNN baseline were trained for 500 epochs with batch size 16 on RTX4090 GPU, with 30 trials per task totaling 300 experiments (Fig.~\ref{fig:FigTaskRollouts}). These ablations functionally approximate representative prior architectures: Workspace Camera Only corresponds to conventional vision-guided suction without tactile feedback~\cite{wang2025demonstratingmultisuctionitempicking}; w/o Peripheral View approximates single-zone camera-based tactile suction systems~\cite{vanVeggel2025, Yuan2025}; and w/o Central View simulates external-camera-only setups~\cite{Lee2024} lacking intimate contact observation. Success is defined as completing the full manipulation sequence without object loss, with primary failure modes: (1) object slipping, (2) search budget exhaustion, and (3) phase transition failures.

\begin{table}[ht] 
\centering
\caption{Task Success Rate Comparison with Statistical Analysis}
\label{tab:success_rates} 
\begin{tabular}{lcc}
\toprule
Configuration & Inclined Transport & Orange Extraction \\
\midrule
\textbf{Ours} & \textbf{73.3\%} & \textbf{66.7\%} \\
w/o Multi-Head Attention & 60.0\% & 53.3\% \\
w/o Peripheral View & 43.3\% & 36.7\% \\
w/o Central View & 46.7\% & 33.3\% \\
Workspace Camera Only & 23.3\% & 0.0\% \\
Baseline & Fail & Fail \\
\bottomrule
\end{tabular}
\end{table}

Table \ref{tab:success_rates} shows the full system achieves 73.3\% (inclined transport) and 66.7\% (orange extraction).
Ablation analysis reveals the central view as critical for contact detection, with its removal reducing performance to 46.7\% and 33.3\% due to loss of tactile sensing.
The peripheral view contributes spatial context essential for approach planning, with its absence degrading performance to 43.3\% and 36.7\%.
Multi-head attention provides 13\% improvements by coordinating dual-zone information during vision-tactile transitions.
Workspace camera alone achieves only 23.3\% and 0.0\% success, confirming that intimate sensing is necessary for contact-critical manipulation.
The BC-RNN baseline failed both tasks (0\% success), frequently becoming stuck or failing to coordinate modality switching---consistent with~\cite{Chi2023}, where recurrent baselines exhibited similar stuck behaviors with multimodal action distributions, confirming the advantage of diffusion policy's action chunking for coordinated suction manipulation.

Failure mode analysis reveals distinct patterns: for configurations lacking peripheral sensing, 60\% of failures stem from phase transition difficulties, while for the full system, 70\% result from physical constraints rather than perceptual limitations, demonstrating that the multimodal architecture effectively addresses perceptual challenges.

\section{Discussion}

\textbf{Object Limitations.} Objects smaller than approximately 15--20~mm fail to establish adequate sealing, practical payload under dynamic manipulation is about 2--3~kg, and highly curved, deeply concave, porous, or rough ($>$1--2~mm feature height) surfaces prevent reliable sealing.
 
\textbf{Wireless Electronics.} While pneumatic connections remain, wireless electronics keep the signal path (CMOS to MCU) onboard and avoid fragile MIPI CSI routing or high-speed slip rings. 
    Although wired embedded solutions (e.g., USB or PoE Ethernet) could provide similar onboard routing, they require platform-specific cable-routing features and tool-flange modifications that reduce portability across robot arms. The wireless design instead enables drop-in deployment on standard tool flanges without mechanical adaptation.
    A 30-minute streaming test shows 27.05 FPS and 43 ms latency, compatible with pneumatic seal dynamics (200--500 ms). 
        Wireless communication is therefore not the bottleneck for the present suction-control loop, though the battery-powered implementation is better suited to intermittent laboratory operation than continuous multi-hour industrial deployment.
 
\textbf{Material and Surface Constraints.} The PDMS membrane showed no observable degradation over 500 cycles, though industrial deployment would still require periodic inspection.

\textbf{Dynamic Performance.} Dynamic stress tests at 3.0 rad/s joint speed and 3.0 rad/s² acceleration maintained stable suction and sensing across varying mass distributions (orange and water-filled bottle; see companion website).
 
\textbf{Sensing and Modality Constraints.} High-frequency switching is limited by CMOS stabilization during illumination changes and is handled through task-level state transitions or learned heuristics. The semi-transparent membrane is also sensitive to ambient lighting, which can occasionally wash out tactile features despite high-intensity internal LEDs.

\textbf{Dual-Mode Complementarity.} Vacuum and Bernoulli modes share the same sensing pipeline but serve complementary domains: vacuum for tactile-rich learning and Bernoulli for contactless handling of delicate objects.

\section{Conclusions}

This paper presented FlexiCup, a multimodal suction cup with wireless electronics and dual-zone vision-tactile sensing through illumination-controlled modality switching. The hardware supports complementary vacuum and Bernoulli actuation while preserving a shared sensing architecture across both modes.

Modular perception-driven grasping achieved 90.0\% (vacuum) and 86.7\% (Bernoulli) success rates, showing that the sensing architecture transfers across both actuation principles; Bernoulli's lower success stems from weaker adhesion rather than sensing failure while still enabling contactless handling of delicate surfaces. End-to-end learning achieved 73.3\% and 66.7\% success on contact-aware manipulation tasks, with multi-head attention improving performance by 13\%. Hardware design files, firmware, and videos are available on the companion website, and future work will target multi-contact tasks, embedded optimization, online mode reconfiguration, and richer tactile feedback.

\end{document}